# Machine learning approach for biopsy-based identification of eosinophilic esophagitis reveals importance of global features

*Tomer Czyzewski[1,#], Nati Daniel[1,#], Mark Rochman[2], Julie M. Caldwell[2], Garrett A. Osswald[2], Margaret H. Collins[3], Marc E. Rothenberg[2], and Yonatan Savir[1,*]*

*Abstract—Goal:* **Eosinophilic esophagitis (EoE) is an allergic inflammatory condition characterized by eosinophil accumulation in the esophageal mucosa. EoE diagnosis includes a manual assessment of eosinophil levels in mucosal biopsies—a time-consuming, laborious task that is difficult to standardize. One of the main challenges in automating this process, like many other biopsy-based diagnostics, is detecting features that are small relative to the size of the biopsy.** *Results:* **In this work, we utilized hematoxylin- and eosin-stained slides from esophageal biopsies from patients with active EoE and control subjects to develop a platform based on a deep convolutional neural network (DCNN) that can classify esophageal biopsies with an accuracy of 85%, sensitivity of 82.5%, and specificity of 87%. Moreover, by combining several downscaling and cropping strategies, we show that some of the features contributing to the correct classification are global rather than specific, local features.** *Conclusions:* **We report the ability of artificial intelligence to identify EoE using computer vision analysis of esophageal biopsy slides. Further, the DCNN features associated with EoE are based on not only local eosinophils but also global histologic changes. Our approach can be used for other conditions that rely on biopsy-based histologic diagnostics.**

*Impact Statement*—**Deep convolutional neural network (DCNN), together with a systematic downscaling approach, can classify esophageal biopsies with high accuracy and reveals a global nature of the histologic features of eosinophilic esophagitis. Our approach of systematic analysis of the image size versus downscaling tradeoff can be used to improve disease classification performance and insight gathering in digital pathology.**

## I. INTRODUCTION

EOSINOPHILIC esophagitis (EoE) is a recently recognized chronic food allergic disease associated with esophageal specific inflammation characterized by high levels of eosinophils [1]. An allergic etiology is strongly supported by the efficacy of food elimination diets, the co-occurrence of EoE with other allergic diseases (e.g., asthma and atopic dermatitis), animal models demonstrating that experimental EoE can be induced by allergen exposure, and the necessity of allergic mediators of inflammation, such as Interleukin 5 and Interleukin 13, on the basis of animal models and clinical studies [1], [2]. Disease pathogenesis is driven by food hypersensitivity and allergic inflammation and multiple genetic and environmental factors [3]. Although a rare disease with a prevalence of approximately 1:2,000 individuals, EoE is now the chief cause of chronic refractory dysphagia in adults and an emerging cause for vomiting, failure to thrive, and abdominal pain in children [1].

Histologically, EoE involves eosinophil-predominant inflammation of the esophageal mucosa. Microscopic examination of esophageal mucosal biopsies is a prerequisite for EoE diagnosis. During esophagogastroduodenoscopy (EGD), several esophageal biopsies are procured. These are then formalin-fixed, embedded, sectioned, and subjected to hematoxylin and eosin (H&E) staining [4], [5]. Subsequently, a pathologist examines the biopsies to determine the peak eosinophil count (PEC) [1], [2], [6] (Fig. 1). In addition to determining PEC, other histopathologic features of EoE include abnormalities of the structural cells, including epithelial cells and fibroblasts comprising the lamina propria. These features can be reliably assessed and quantified using the newly developed EoE Histology Scoring System (HSS) [7]. This system not only reports the presence or absence of the features but also takes into account grade (severity) and stage (extent). This scoring system is trainable across pathologists [7]. However, considerable disagreement can occur among certain observers, at least based on PEC [8], and even for trained observers, scoring esophageal biopsies requires a non-trivial time input.

During the last few years, deep learning and, in particular, deep convolutional neural networks (DCNNs) have become a significant component of computer vision. Unlike classical machine learning techniques, deep learning involves the net performing representation learning, which allows the machine to be fed raw data and to discover the representations needed for detection or classification automatically [9]–[12]. In particular, deep learning is used for the classification and diagnosis of conditions in which the diagnosis is based on histomorphology, such as cancer [12], [13]. However, the application of deep learning to medical applications poses two unique challenges: first, DCNN training requires a large number of images (hundreds to millions); and second, the size of the relevant objects within the images is small [14], [15].

[1]Dept. of Physiology, Biophysics and System Biology, Faculty of Medicine, Technion, Haifa, 35254, Israel. [2]Division of Allergy and Immunology, [3]Division of Pathology, Cincinnati Children's Hospital Medical Center, Department of Pediatrics, University of Cincinnati College of Medicine, Cincinnati, OH 45229-3026, USA.

[#]These Authors contributed equally to this work.
[*]Correspondence: yoni.savir@technion.ac.il

Here, we developed a method based on DCNN and downscaling of esophageal biopsy images at different frequencies. By comparing the results of each frequency, we aimed to deduce whether the scattering is global (i.e., features appear diffusely throughout the tissue image) or local (i.e., features appear in only specific and/or discrete locations within the image). We developed a classifier that distinguishes between images of H&E-stained esophageal biopsies from patients with active EoE and non-EoE control patients with high accuracy. We show that some of the features that underlie the correct classification of disease are global in nature.

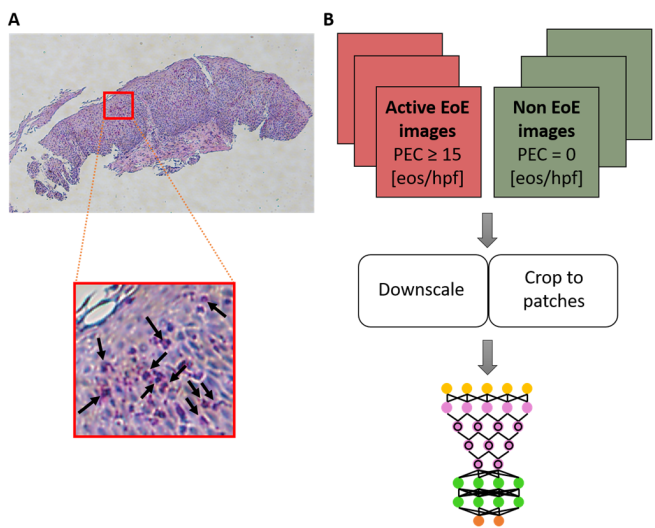

Fig. 1. (A) Example of a full-size hematoxylin and eosin (H&E)-stained esophageal biopsy slide from a patient with active eosinophilic esophagitis (EoE). The red square marks an example of an area containing eosinophils (bright pink cells with purple nuclei; several examples are indicated by black arrows in the inset). (B) Schematics of the platform. Images (magnification 80X) of research slides (from one esophageal research biopsy per patient) are labeled as EoE or non-EoE on the basis of a pathologist's analysis of corresponding clinical slides associated with the same endoscopy during which the research biopsy was obtained. The full-size images are downscaled and/or cropped using various approaches to smaller images that are then used to train a deep convolutional neural network (DCNN). eos, eosinophils; hpf, high-power field; PEC, peak eosinophil count.

## II. MATERIALS AND METHODS

### A. Dataset

This study was performed under the Cincinnati Children's Hospital Medical Center (CCHMC) IRB protocol 2008-0090. Subjects undergoing endoscopy (EGD) for standard-of-care purposes agreed to donate additional gastrointestinal tissue biopsies for research purposes and to have their clinical, histologic, and demographic information stored in a private research database. One distal esophageal biopsy per patient was placed in 10% formalin; the tissue was then processed and embedded in paraffin. Sections (4 μm) were mounted on glass slides and subjected to H&E staining, in a manner identical to the preparation of standard-of-care biopsies. Biopsies were viewed at 80X magnification using the Olympus BX51 microscope, and one photograph of each biopsy was taken using the DP71 camera. Images were classified into categories on the basis of the clinical pathology report associated with the distal esophagus biopsies that were obtained for clinical analysis during the same endoscopy during which the biopsy for research purposes was procured. In this study, we used images defined as being derived from individuals with active EoE (biopsy with PEC ≥15 eosinophils [eos]/400X high-power field [hpf]) or from non-EoE control individuals (biopsy with PEC = 0 eos/hpf); (n = 210 non-EoE; n = 210 active EoE). The images were taken with digital microscopy at different resolutions: 4140X3096 pixels, 2010X1548 pixels, or 1360X1024 pixels. In the original dataset, the number of images per category and at each resolution was not equal. Therefore, to avoid training bias, the images were randomly selected to build non-biased training and validation sets. In this new dataset, the number of images in each category was equal (training set: n = 147 active EoE, n = 147 non-EoE; validation set: n = 63 active EoE, n = 63 non-EoE). Additionally, the number of images per resolution was equal in each category (4140X3096 resolution: n = 29; 2010X1548 resolution: n = 126; 1360X1024 resolution: n = 55).

### B. Downscale approaches and training

Two methods were employed to address the challenge of training on high-resolution images containing small features: first, downscaling the original image with the potential of losing the information associated with small features [14]; and second, dividing the images into smaller patches and analyzing each of the patches [16]. Although the second approach solves the image size challenge, if the relevant small feature (e.g., a local increase in eosinophil density) appears in only a few patches, many patches that do not contain the small feature are still labeled as positive. As a result, the false-positive prediction might significantly bias the final diagnosis. Yet, this method indicates whether the scatter of the features is global or local by carefully comparing it to a random classifier.

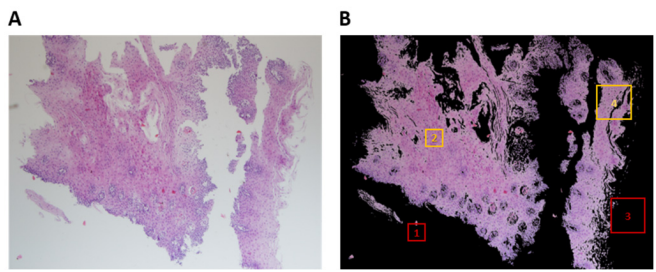

Fig. 2. Steps in processing esophageal biopsy images to produce patches. (A) A typical image of a hematoxylin and eosin (H&E)-stained esophageal biopsy section obtained from an individual with active EoE. The image was taken at 80X magnification. (B) The same image after background removal with an illustration of tissue coverage criteria per patch size to meet the threshold for inclusion in training or validation sets. Box 1 (red): patch of 224X224 pixels with less than 10% tissue coverage. Box 2 (yellow): patch of 224X224 pixels with greater than 10% tissue coverage. Box 3 (red): patch of 448X448 pixels with less than 10% tissue coverage. Box 4 (yellow): patch of 448X448 pixels with greater than 10% tissue coverage.

In this work, the chosen DCNN was ResNet50. Four different DCNNs were trained, wherein each of the input image sizes was obtained differently: 1) cropping the full image to patches of 224X224 pixels (the optimal size for ResNet50), 2) cropping the full image to patches of 448X448 pixels and downscaling them to 224X224, 3) downscaling the original image to 224X224 pixels resolution, and 4) downscaling the original image to 1000X1000 pixels resolution (Table I). This resolution was chosen because it represents nearly the maximum resolution possible for training on Nvidia 1080TI with a minimal mini-batch size of four images. Downscaling was done using bicubic interpolation.

Patches were cropped with a sliding window of the desired input (224X224, 448X448 pixels) with steps of half of the input resolution for overlay, covering the full original images (an example of a full image is shown in Fig. 2A). Subsequently, only patches that had more than 10% tissue comprising the patch were chosen for training and validation sets (Fig. 2B). All valid patches were used for training. During training, rotation, translation, scaling, and flipping augmentation were performed.

## III. RESULTS

Table I summarizes the whole image classification results for the four downscale and/or crop approaches employed. First, we downscaled the original images to two different input image resolutions. If the majority of the information that defines the condition were local, we would expect that downscaling, resulting in smooth local features, would have a significant effect on the classification quality. Surprisingly, we found that downscaling the original images to a size of 1000X1000 did not result in a random classification, but instead resulted in a true positive rate (TPR) of 74.6% and a true negative rate (TNR) of 96.8%. These results suggest that some of the information that defines the condition is local but is large enough to sustain the downscaling; alternatively, the information could be global. The bias towards negative classification (predicted prevalence [PP] <0.5), as indicated by the PP of 0.39, suggests that the information that determines the condition is more local, leading to more positive-labeled images having the same feature as negative-labeled images. Downscaling the full images even further to a size of 224X224 reduced both the TPR and the TNR. Yet, consistent with the hypothesis that the information that defines the positive images is more sensitive to downscaling, the PP remained similar, and the TPR was reduced more than the TNR (Δ9.5% and Δ7.9%, respectively].

Next, we classified the whole images according to the sub-classification of their patches. The predicted label assigned to the whole image (i.e., active EoE or non-EoE) resulted from the majority vote of the predicted labels of its patches (i.e., if ≥50% of patches were classified as active EoE, the whole image was classified as active EoE; if ≥50% of patches were classified as non-EoE, the whole image was classified as non-EoE). First, each image was parsed into patches, each with a size of

TABLE I
WHOLE IMAGE PREDICTION

| Original Image | Final DCNN input image resolution | Active EoE (TPR) | Non-EoE (TNR) | Accuracy | Predicted Prevalence (PP) |
|---|---|---|---|---|---|
| Full Image | 1000x1000 (Downscale) | 74.6% | 96.8% | 85.7% | 0.39 |
| Full Image | 224x224 (Downscale) | 65.1% | 88.9% | 77.0% | 0.38 |
| Patch = 448x448 | 224x224 (Downscale) | 82.5% | 87.3% | 84.9% | 0.48 |
| Patch = 224x224 | 224x224 | 82.5% | 77.8% | 80.2% | 0.52 |

Whole image classification results for four downscale and/or crop approaches. The validation cohort of images (n = 63 active EoE; n = 63 non-EoE) was the same for each of the classifiers. True positive rate (TPR; number of images classified as active EoE / number of active EoE images x 100), true negative rate (TNR; number of images classified as non-EoE / number of non-EoE images x 100), accuracy (number of images accurately classified as either active EoE or non-EoE / total number of images x 100), and predicted prevalence (total number of images classified as active [i.e., true positive + false positive number of images] / total number of images) for each method are shown. DCNN, deep convolutional neural network.

448X448 that were then each downscaled to a size of 224X224. In this case, no substantial classification bias resulted; the PP of 0.48 and the TPR of 82.5% increased substantially compared to the two downscaling methods described previously (Table I).

Using patches of 224X224 that did not undergo downscaling yielded a similar TPR of 82.5%; however, the TNR decreased to 77.8%. This is likely due to the inherent tradeoff between the local and global information contained within the images. If an image is larger, it contains more global information, but the downscaling that is required prior to its input into the net is larger; thus, small features are smoothed out to a greater degree. In our case, using a 448X448 patch with downscaling provided a better TNR of 87.3% than did using smaller patches of 224X224 without downscaling. Figure 3 summarizes the effect of the initial patch size and downscaling factor in the receiver operating characteristic (ROC) space.

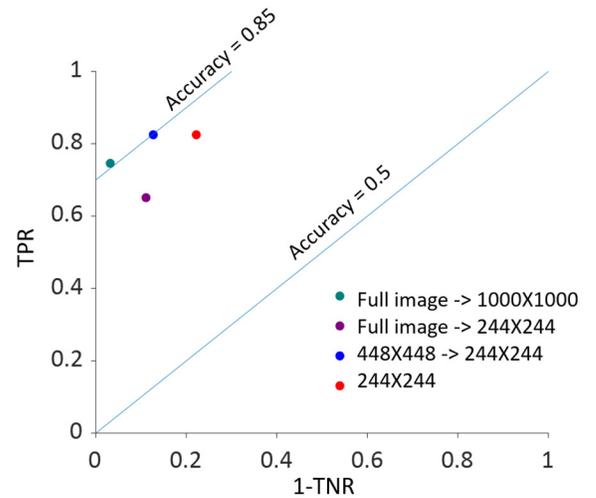

Fig. 3. Classification results as a function of initial image size and downscaling factor in the receiver operating characteristic (ROC) space. For each of the four downscale and/or crop approaches utilized to analyze the validation cohort of images (n = 63 active EoE; n = 63 non-EoE), the true positive rate (TPR) vs. (1 - the true negative rate [TNR]) with TPR and TNR expressed as proportions is graphed. Blue lines highlight accuracy measurements of 50% and 85% expressed as proportions.

To further analyze the tradeoff between locality and downscale factor, we evaluated the classification performance of the patches themselves (Table II). The results are consistent with the whole image majority vote classification. In particular, both the TNR of 79.7% and TPR of 77.0% of the 448X448 patch downscaled to 224x224 are higher than those of the non-scaled 224X224 patch. These results indicate that incorporating more information in a patch is more important than

TABLE II
PATCH PREDICTION

| Original image | Final DCNN input image resolution | Active EoE (TPR) | Non-EoE (TNR) | Accuracy | Predicted Prevalence (PP) |
|---|---|---|---|---|---|
| Patch = 448x448 | 224x224 (Downscale) | 77.0% | 79.7% | 78.3% | 0.49 |
| Patch = 224x224 | 224x224 | 73.3% | 75.2% | 74.2% | 0.49 |

Classification results for individual patches. The validation cohort of images (n = 63 active EoE; n = 63 non-EoE) was subjected to cropping into patches with the indicated pixel sizes and downscaled when indicated. True positive rate (TPR; number of patches classified as active EoE / number of active EoE patches x 100), true negative rate (TNR; number of patches classified as non-EoE / number of non-EoE patches x 100), accuracy (number of patches accurately classified as either active EoE or non-EoE / total number of patches x 100), and predicted prevalence (total number of images classified as active [i.e., true positive + false positive number of images] / total number of images) for each patch size and downscaling method (if applicable) are shown. DCNN, deep convolutional neural network; TPR, true positive rate; TNR, true negative rate.

downscaling by a factor of two and supports the notion that global information drives the classification for EoE.

To determine the effect of locality on the classification, we compared the distribution of prediction probability for patches with a size of 224X224 that did not undergo downscaling in two cases. In the first, each patch was labeled with the same label as the original image from which it was derived. In the second, each patch was assigned a random label.

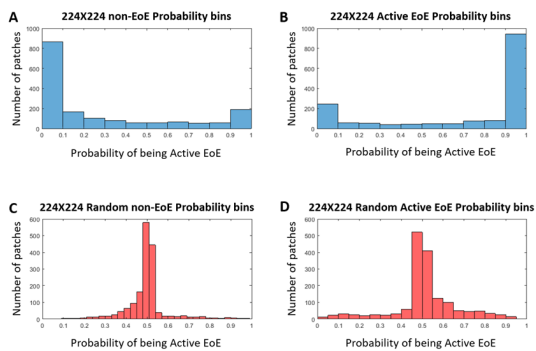

Fig. 4. Prediction ability of nonrandom (blue) and random (red) classifier. (A) 224X224: histogram of the number of patches derived from non-EoE images vs. the probability that they will be classified as active EoE by the nonrandom classifier. (B) 224X224: histogram of the number of patches derived from active EoE images vs. the probability that they will be classified as active EoE by the nonrandom classifier. (C) Random 224X224: histogram of the number of patches derived from non-EoE–labeled images vs. the probability that they will be classified as active EoE by the random classifier. (D) Random 224X224: histogram of the number of patches derived from active EoE–labeled images vs. the probability that they will be classified as active EoE by the random classifier.

Figure 4 shows the distribution for each case. In the case in which the patch labels are true (Fig. 4A, B), the distribution is bi-modal. In the case in which the patch labels are random (Fig. 4C, D), most of the patches are ambiguous, and thus the distribution is unimodal around 0.5. These collective case findings suggest that most of the patches that are classified correctly are not ambiguous. This indicates that the local patch labeling carries information that is relevant for the majority of the patches.

## IV. DISCUSSION

One of the main challenges in digital pathology is that the features of the conditions are very small compared with the size of the sample. This feature-sample size disparity leads to an inherent tradeoff between the size of the analyzed image and the downscaling factor. In the case of small, local features, visualizing the image as smaller patches may impede the classification because most of the patches will not include the small, local features. However, if local features are the primary source of information about the condition, downscaling the whole image may smooth them out.

Herein, we used DCNN and different downscaling and/or cropping approaches to achieve ~85% accuracy in distinguishing active EoE from non-EoE esophageal biopsies, despite the relatively small number of labeled images utilized for training (n = 147 active EoE and n = 147 non-EoE). Although labeling relied primarily on a local feature (PEC ≥15 eos/hpf), our results support that EoE is also associated with additional global histopathologic features that are learned by the classifier. Figure 5 illustrates possible scatter patterns for features that contribute to disease diagnosis. Of note, the features could be clustered locally (e.g., a local increase in density of eosinophils), or they could be distributed uniformly throughout the tissue (e.g., morphology of structural cells comprising the tissue).

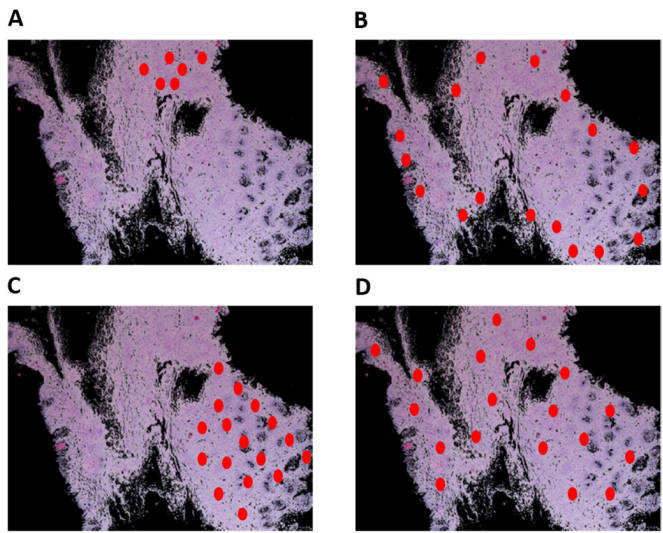

Fig. 5. Schematic of various potential distributions of local patterns within an esophageal biopsy section. An esophageal biopsy image is shown; red ovals denote a local feature that contributes to disease diagnosis. (A) Local pattern confined to a specific place in the tissue. (B) Local pattern distributed at the edge of the tissue. (C) Local pattern restricted to only half of the tissue. (D) Global pattern spread all over the tissue.

The fact that images that were cropped into patches but were downscaled by a factor of greater than 10 (in terms of the number of pixels) provided low TPR, suggests that the features associated with the condition were not big enough for the classification task. However, if the features were distributed only locally (e.g., Fig. 5A-C), many patches cropped from the whole image would not include the features, and thus the classification according to patches would fail. However, in this study of EoE, most of these cropped patches were labeled correctly. Moreover, the classification was better with 448X448 patches downscaled to 224X224 than non-scaled 224X224 patches, suggesting presence of global features (Fig. 5D).

Our results thus indicate that although the original labeling was based primarily on local features, additional global features are associated with EoE (Fig. 5D). This global information allows a classification with minimal PP bias (PP 0.49) and with only a small number of images. Our work highlights the importance of systematic analysis of the image size vs. downscaling tradeoff, particularly in digital pathology, for improving classification and gaining insight into the features' spatial distribution underlying a condition. These findings present an initial artificial intelligence approach to diagnosing EoE using digital microscopy and have implications for analyzing other biopsy-based disease diagnoses.

ACKNOWLEDGMENT

Y.S. was supported by the American Federation for Aging Research, Israel Science Foundation #1619/20, Rappaport Foundation, The Prince Center for Neurodegenerative Disorders of the Brain #828931. M.E.R. was supported by NIH R01 AI045898-21, the CURED Foundation and Dave and Denise Bunning Sunshine Foundation. The authors would like to thank Tanya Wasserman for valuable discussions and Shawna Hottinger for editorial support.